\documentclass[letterpaper, 10 pt, conference]{ieeeconf}  

\IEEEoverridecommandlockouts                              

\usepackage{amsmath} 
\usepackage{amssymb}  
\usepackage{graphicx}

\usepackage{algorithm}
\usepackage{algpseudocode}

\usepackage[hidelinks]{hyperref}
\usepackage{tcolorbox}

\usepackage{float}

\usepackage{cite}

\usepackage{array}

\usepackage{booktabs}

\title{\LARGE \bf
From Words to Wheels: Automated Style-Customized Policy Generation for Autonomous Driving
}

\author{Xu Han$^{1}$, Xianda Chen$^{2}$, Zhenghan Cai$^{3}$, Pinlong Cai$^{4}$, Meixin Zhu$^{2}$, Xiaowen Chu$^{1*}$
\thanks{*Corresponding author: Xiaowen Chu, xwchu@ust.hk}%
\thanks{$^{1}$Data Science and Analytics Thrust, Information Hub, The Hong Kong University of Science and Technology (Guangzhou), Guangzhou, China}%
\thanks{$^{2}$Intelligent Transportation Thrust, Systems Hub, The Hong Kong University of Science and Technology (Guangzhou), Guangzhou, China}%
\thanks{$^{3}$Department of Automation, School of Electrical Automation and Information Engineering,Tianjin University, Tianjin, China}%
\thanks{$^{4}$Shanghai Artificial Intelligence Laboratory, Shanghai, China}%
\thanks{This is a preprint version.}%
}

\begin{document}

\markboth{}%
{Shell \MakeLowercase{\textit{et al.}}: A Sample Article Using IEEEtran.cls for IEEE Journals}
\maketitle

\begin{abstract}

Autonomous driving technology has witnessed rapid advancements, with foundation models improving interactivity and user experiences. However, current autonomous vehicles (AVs) face significant limitations in delivering command-based driving styles. Most existing methods either rely on predefined driving styles that require expert input or use data-driven techniques like Inverse Reinforcement Learning to extract styles from driving data. These approaches, though effective in some cases, face challenges: difficulty obtaining specific driving data for style matching (e.g., in Robotaxis), inability to align driving style metrics with user preferences, and limitations to pre-existing styles, restricting customization and generalization to new commands. This paper introduces Words2Wheels, a framework that automatically generates customized driving policies based on natural language user commands. Words2Wheels employs a Style-Customized Reward Function to generate a Style-Customized Driving Policy without relying on prior driving data. By leveraging large language models and a Driving Style Database, the framework efficiently retrieves, adapts, and generalizes driving styles. A Statistical Evaluation module ensures alignment with user preferences. Experimental results demonstrate that Words2Wheels outperforms existing methods in accuracy, generalization, and adaptability, offering a novel solution for customized AV driving behavior. Code and demo available at \url{https://yokhon.github.io/Words2Wheels/}.

\end{abstract}


\section{INTRODUCTION} \label{section:intro}


Autonomous driving technology has advanced rapidly, especially with the rise of foundation models that improve interactivity and user experience\cite{cui2024survey,zhou2024vision,wen2023dilu,hazra2024revolve}. Despite these advancements, significant room for improvement remains, particularly regarding user experience. One key criticism is the uniform driving style of current autonomous vehicles (AVs)\cite{nativel2023exploration,jing2020determinants,liao2024review}. For example, 52.46\% of AVs involved in crashes were rear-ended, 1.6 times the rate of conventional vehicles, highlighting the need for gentler braking styles\cite{liu2021crash}.

\begin{figure}[t]
\centering
\includegraphics[width=\linewidth]{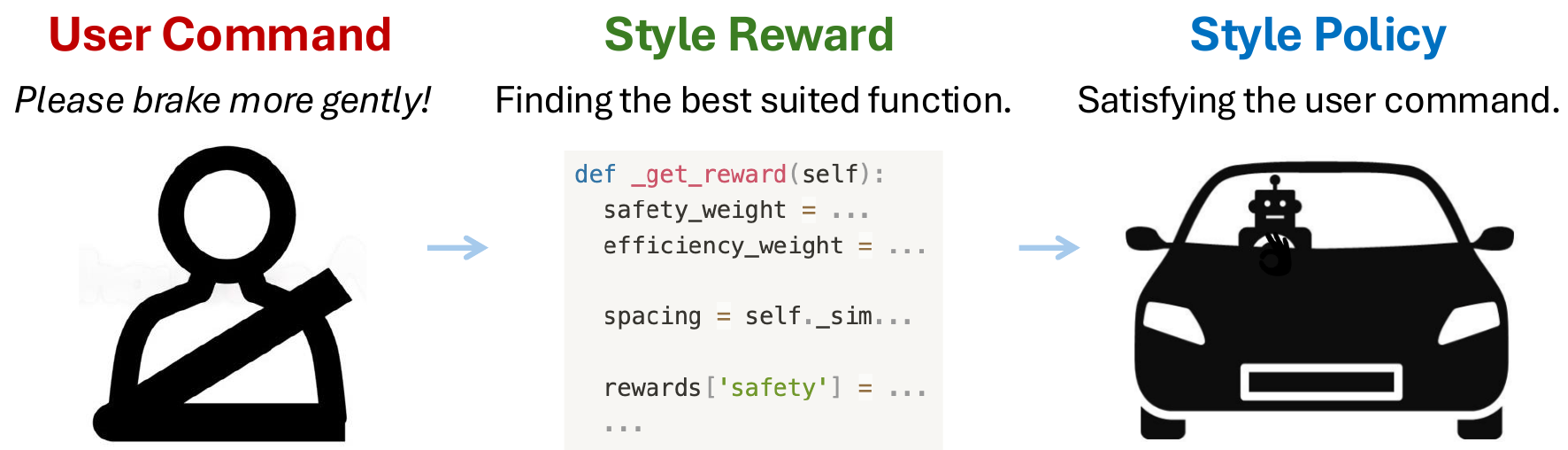}
\caption{Words2Wheels automatically generates customized driving policies in alignment with user commands by LLM-powered reward design.}
\label{intro}
\end{figure}


Driving style, or the characteristic way a vehicle operates, is crucial for autonomous driving\cite{mantouka2022deep}. While many studies have explored driving styles, early approaches relied heavily on classification or calibration of predefined styles or driving patterns\cite{martinez2017driving,chu2017curve,suzdaleva2019two,wang2017driving,chen2024aggfollower,vaitkus2014driving,murphey2009driver}. For instance, Wang et al. used a hidden semi-Markov model to extract driving patterns\cite{wang2018driving}, and Deng et al. applied a hidden Markov model to braking analysis\cite{deng2017driving}. However, these manually designed styles depend heavily on expert input, making the process cumbersome and subjective, requiring frequent adjustments\cite{huang2021driving}.

\begin{figure*}[t!]
\centering
\includegraphics[width=\textwidth]{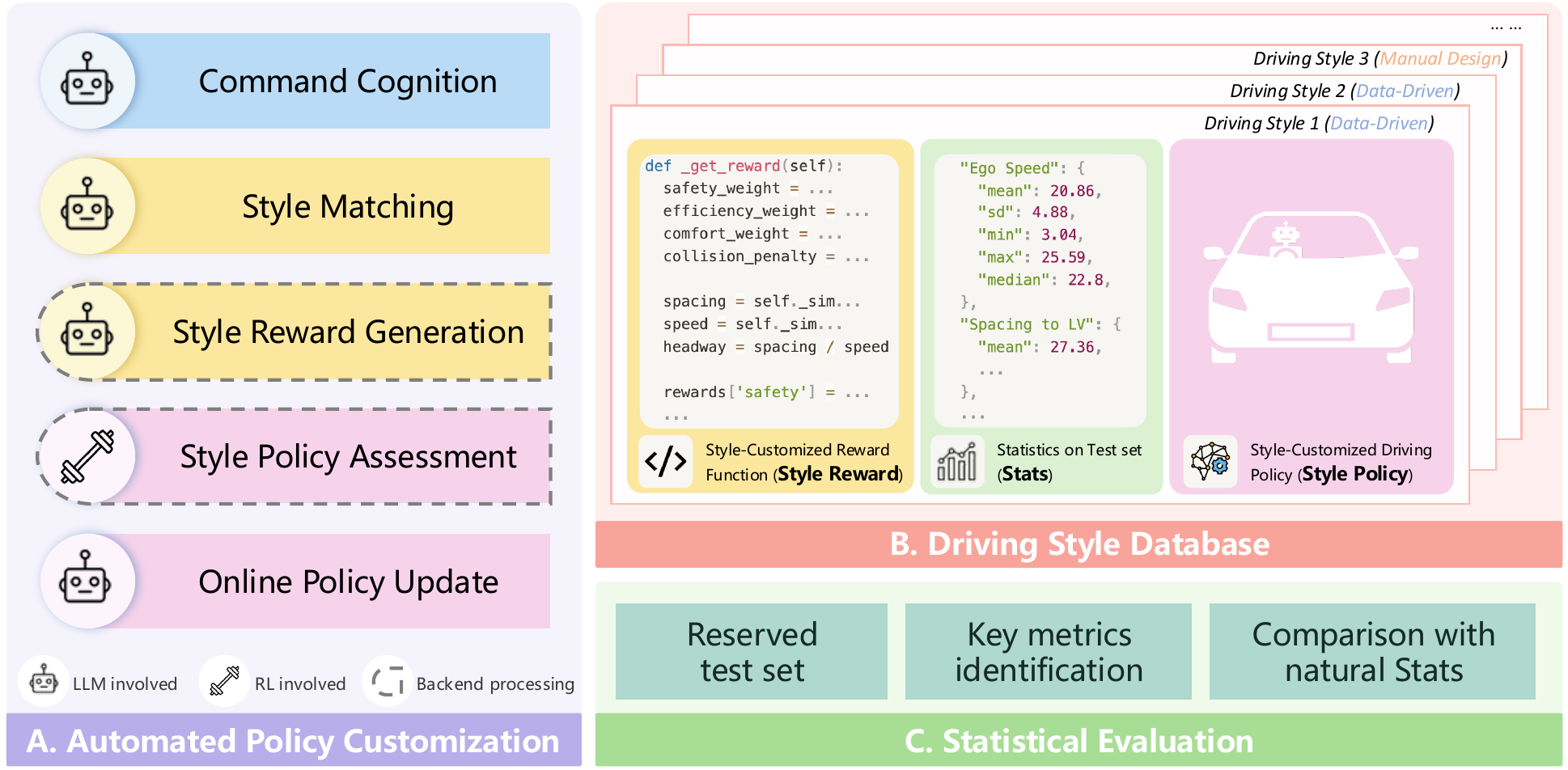}
\caption{(A) \textbf{Workflow of Words2Wheels}: When a natural language command is received, the system matches it with a style from the database. Style Reward generation and policy training run simultaneously in the backend, resulting in a new Style Policy that may outperform the existing one and replace it. (B) \textbf{Driving Style Database}: This repository stores Style Rewards (initially from both data-driven and human-designed methods), Style Policies, and their statistics. It manages the increasing variety of driving styles and supports the automated policy customization. (C) \textbf{Statistical Evaluation Module}: This module ensures that the generated driving styles closely align with user commands by evaluating them against natural driving behaviors.}
\label{framework}
\end{figure*}

In recent years, data-driven approaches have gained traction, using methods such as clustering and Inverse Reinforcement Learning (IRL) to derive driving policies with varying styles from natural driving behavior data\cite{chu2023review,kuderer2015learning,zhou2020modeling,wang2016rapid,chen2024editfollower,constantinescu2010driving}. Zhao et al. uses model-based IRL to learn and adapt to individual driver preferences from historical data\cite{zhao2022personalized}. Rosbach et al. applied maximum entropy IRL to automatically tune reward functions based on human driving demonstrations\cite{rosbach2019driving}. Nevertheless, these approaches face several challenges when applied to autonomous driving. \emph{Firstly, they require a small amount of driving behavior data to match with a specific driving style, but obtaining such data is particularly difficult in Robotaxi scenarios, leading to issues with style matching}\cite{zhao2022personalized}. \emph{Secondly, the metrics used to quantify driving styles are diverse, and the inability to prioritize these metrics according to user preferences can severely restrict the effectiveness of style matching. Lastly, these methods are limited to selecting from existing styles and cannot customize a style based on user needs, potentially leading to situations where no existing style aligns with the user command, known as the generalization problem}\cite{liu2022inverse,sheng2022study}. These challenges limit the applicability of these methods in autonomous driving.

More recently, foundation models have been explored to process user instructions\cite{chen2024driving,huang2024drivlme,yang2024human,wang2024omnidrive,xu2024drivegpt4}. Cui et al. integrated large language models (LLMs) to personalize driving experiences through dynamic interaction\cite{cui2023receive}, while Yang et al. aligned LLM-powered agents with human driving styles from interviews\cite{yang2024driving}. While these methods can adjust specific behaviors like speed and lane changes based on natural language input, they fall short in adjusting the overall driving policy at the style level\cite{cui2023receive,chen2024driving}. Additionally, these approaches often involve the foundation model in most decision loops, raising concerns about reliability and efficiency\cite{wang2023drivemlm,ma2024lampilot}.

To address these challenges, we propose Words2Wheels, a framework for automatically customizing driving policies based on user commands. The core innovation is the use of a Style-Customized Reward Function (\textbf{Style Reward}) to bridge user input and a Style-Customized Driving Policy (\textbf{Style Policy}), as shown in Fig. \ref{intro}. In Reinforcement Learning (RL), the reward function guides the driving policy to optimize long-term rewards and serves as a policy index\cite{sutton1999reinforcement}. These Style Rewards are less complex than user commands or traffic scenarios, enabling efficient retrieval of Style Policy\cite{han2024generating}. By leveraging LLMs and Retrieval-Augmented Generation (RAG) technology\cite{lewis2020retrieval}, Words2Wheels precisely matches driving styles using quantitative metrics. With LLMs’ ``zero-shot'' generation capabilities\footnote{Zero-shot refers to no need for user driving data here. It’s few-shot in generating reward functions based on existing Style Rewards.}, Words2Wheels can generalize to new commands and styles without user behavior data. The framework is also extensible, allowing for features like fuzzy memory via LLMs.


To ensure the Style Policy matches user commands, we developed two key modules. First, the \textbf{Driving Style Database}, a repository containing Style Rewards, Style Policies, and statistical features such as jerk during braking, as shown in Fig. \ref{framework}. These features guide the LLM's analysis, enabling fast selection of appropriate Style Policies and generation of new Style Rewards when needed. New Style Policies that outperform existing ones can replace them, with all updates added to the database, continuously expanding the style range. Second, the \textbf{Statistical Evaluation} module measures how well the Style Policy aligns with the user command. Using a reserved test dataset, this module prevents overfitting and provides benchmarks for evaluating policies. It automatically selects relevant metrics, compares the policies with natural data, and offers a reliable evaluation of driving style.



To the best of our knowledge, Words2Wheels is the first framework that automatically customizes style-aware driving policies to align with natural language user commands. The primary contributions of this paper include:

\begin{itemize}
    \item Developing the Words2Wheels framework, which automatically customizes driving policies to bridge the gap between user preferences and driving behavior.
    \item Introducing a conceptual Driving Style Database that enables efficient retrieval and continuous enrichment of driving styles based on user commands.
    \item Designing a Statistical Evaluation module to automatically and accurately measure the alignment between driving policies and user commands.
    \item Conducting comprehensive experiments to validate the effectiveness of Words2Wheels, demonstrating its capabilities in accuracy, generalization, and reliability.
\end{itemize}

\section{PROBLEM FORMULATION}

\subsection{Markov Decision Process}




In autonomous driving, decision-making is often modeled as a Markov Decision Process (MDP)\cite{zhu2018human,han2023ensemblefollower}. An MDP consists of states \(S\) representing possible environment configurations, actions \(A\) available to the vehicle, transition probabilities \(P(s'|s, a)\) for moving between states, a reward function \(R(s, a)\) that provides feedback for actions, and a discount factor \(\gamma\) that balances future rewards. The goal is to find a policy \(\pi\) that maximizes the expected cumulative reward, \(E \left[ \sum_{t=0}^\infty \gamma^t R(s_t, a_t) \right]\).


Reinforcement Learning (RL) is employed to identify the optimal policy \(\pi^*\), which maximizes this reward. The agent learns by interacting with the environment—observing states, taking actions, and receiving rewards. The optimal value function \(V^*(s)\), representing the expected cumulative reward from state \(s\), satisfies the Bellman equation\cite{sutton1999reinforcement}: 

\[
V^*(s) = \max_{a \in A} \left[ R(s, a) + \gamma \sum_{s' \in S} P(s'|s, a) V^*(s') \right]
\]





IRL focuses on recovering the reward function \(R(s, a)\) from expert demonstrations\cite{ziebart2008maximum}, inferring the reward that makes the observed policy optimal. This enables the vehicle to replicate human-like decision-making.

Both RL and IRL are essential for developing autonomous systems, enabling them to learn from simulations and real-world data\cite{you2019advanced,chen2023follownet}.

\subsection{Driving Style Alignment}

In autonomous driving, users often express their preferences for driving styles in natural language. The task is to generate a driving policy \(\pi_{style}\) that aligns with a user-specified driving style, expressed as a command \(C\).

The Driving Style Alignment problem is formulated as:

\begin{enumerate}
    \item User Command Interpretation: Given a natural language command \(C\) that describes the desired driving style, the system must interpret the command and map it to a driving style representation \(\theta_{style} \in \Theta\). This mapping is expressed as:

    \[
    \mathcal{F}: \mathcal{C} \rightarrow \Theta
    \]
    
    where \(C \in \mathcal{C}\) is the user command, and \(\mathcal{F}(C) = \theta_{style}\) is the inferred driving style representation, which indicates the reward function \(R(s, a)\) in this work.

    \item Driving Policy Generation and Optimization: Based on the interpreted driving style \(\theta_{style}\), the system generates a driving policy \(\pi_{style}\) that minimizes the deviation from the user’s expected behavior, described by \(C\). This step can be expressed as:

    \[
    \pi_{style}^* = \mathcal{G}(\theta_{style}) = \arg\min_{\pi \in \Pi} d(\pi, C)
    \]
    
    where \(\mathcal{G}: \Theta \rightarrow \Pi\) maps the driving style \(\theta_{style}\) to the optimal policy \(\pi_{style}^* \in \Pi\), minimizing the deviation \(d(\pi, C)\) from the expected behavior. This mapping \(\mathcal{G}\) is conducted by RL in the Words2Wheels framework.
\end{enumerate}






\section{Methodology}\label{section:approach}

\subsection{Automated Generation}

\begin{figure}[h]
    \centering
    \includegraphics[width=\linewidth]{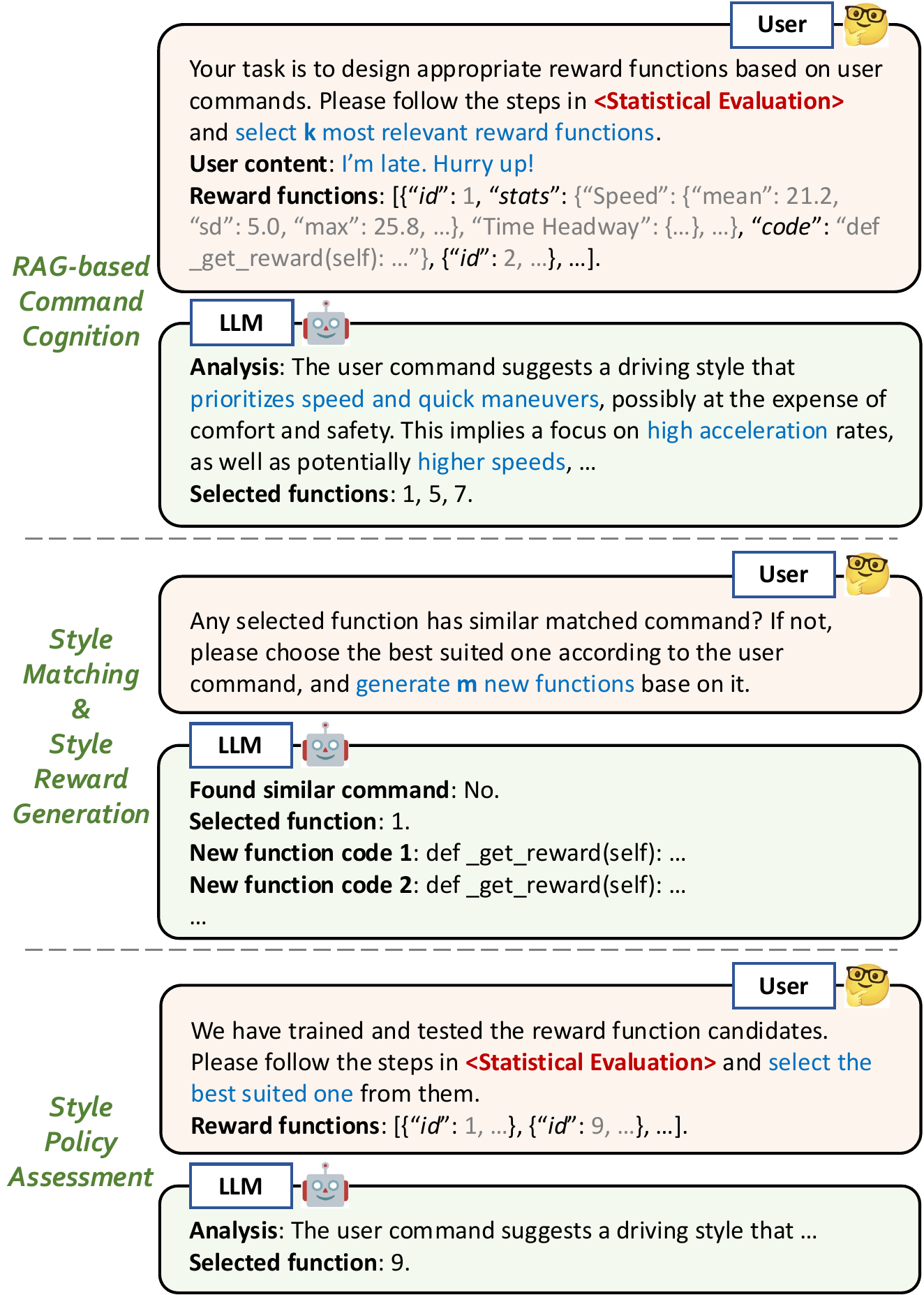}
    \caption{Example of automated Style Policy generation.}
    \label{fig:conv_1}
\end{figure}

Upon receiving a user command like ``I'm late. Hurry up!'', the Words2Wheels system automatically generates a driving policy aligned with the directive. As shown in Fig. \ref{fig:conv_1}, the process starts by applying RAG principles, where the LLM selects the top $k$ relevant style rewards from the Driving Style Database, improving focus and task performance. This step can be parallelized for efficiency.

To ensure reliable selection, the LLM consults the Statistical Evaluation module. If a selected Style Reward has matched a semantically similar command, Words2Wheels directly uses its Style Policy (the fuzzy memory function), enhancing efficiency. If no match is found, the LLM generates $m$ new reward functions based on the most relevant Style Reward. The LLM's generation goes beyond simple parameter adjustments and may involve logic modifications or restructuring to handle generalization challenges effectively.


The initially selected Style Reward provides a temporary solution since its policy is already trained. Meanwhile, the newly generated reward functions produce new Style Policies via RL, where the trainings can be parallelized. Finally, using the Statistical Evaluation module, Words2Wheels determines if the new policies outperform the temporary one, delivering the best match to the user command. This process is fully automated, reducing biases from subjective judgments.

\subsection{Driving Style Database}


The Driving Style Database is a key component of Words2Wheels. As shown in Fig. \ref{framework}, each record includes a Style Reward, a Style Policy, and analytical data. Style Rewards are programming codes, while Style Policies are stored as pre-trained neural networks. Analytical data, generated by the Statistical Evaluation module, can be saved in JSON format. These elements can be embedded as high-dimensional vectors to improve LLM's retrieval efficiency.



The database allows the LLM to select existing styles or use reward functions as templates to generate new ones. It is beneficial to populate the database with pre-existing reward functions at the start, which can enhance the quality of the LLM's generations. Existing research on driving reward design provides valuable references, ranging from human-designed rewards\cite{zhu2020safe, abouelazm2024review} to data-driven rewards derived from real-world data using clustering and IRL\cite{zhao2022personalized, qiu2024driving}.


The greater the variety of reward functions, the more examples the LLM can reference. As Words2Wheels operates, new driving styles generated by the system expand the database, creating a broader range of styles and reducing reliance on RL, improving overall efficiency. User commands are also stored, enabling the fuzzy memory function.

\subsection{Statistical Evaluation}

\begin{figure}[h]
    \centering
    \includegraphics[width=\linewidth]{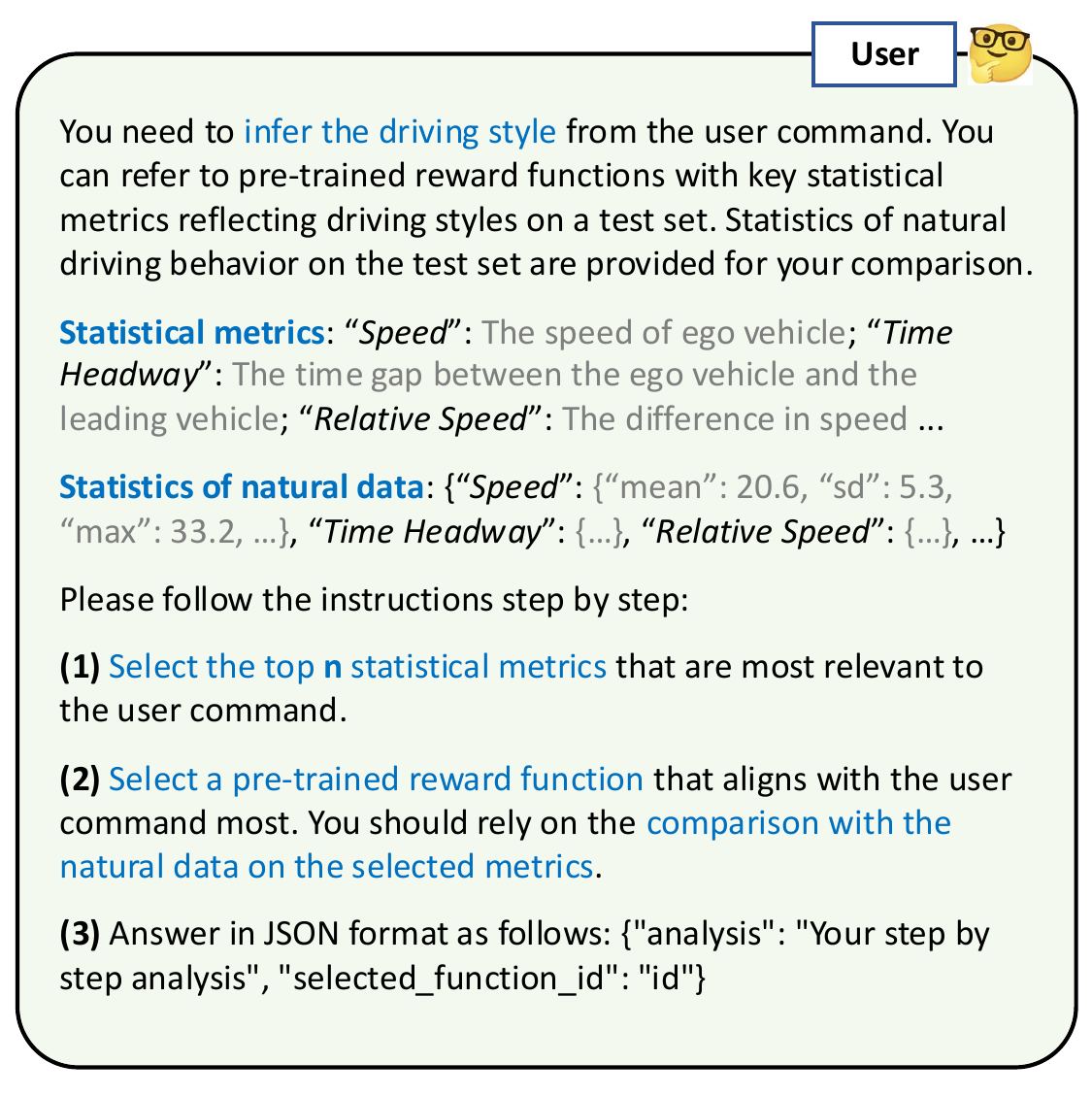}
    \caption{Simplified example of Statistical Evaluation module.}
    \label{fig:conv_2}
\end{figure}


The Statistical Evaluation module is essential for generating statistical data on driving behavior and helping the LLM accurately assess how well a Style Policy aligns with a user command. To ensure reliability, we reserve a test dataset to simulate driving behavior and collect relevant data. From this, various metrics are calculated, such as speed, acceleration, jerk, spacing, and time headway, as informed by prior research on driving styles\cite{sagberg2015review,meiring2015review,martinez2017driving}.



These metrics help the LLM evaluate Style Policies by comparing the statistics to the user command, as shown in Fig. \ref{fig:conv_2}. The prompt includes a brief introduction to the metrics and natural driving behavior statistics from the test set to provide a baseline. Using a Chain-of-Thought approach\cite{wei2022chain}, the LLM selects the $n$ most relevant metrics and compares the policies to natural driving data, enabling objective style evaluation.

Customizing the test set allows for expanded functionality. For instance, setting the test set to a specific Style Policy’s driving data enables fine-tuning based on that policy. Additionally, spatio-temporal filtering of test set data can achieve more precise quantitative analyses.


\section{EXPERIMENTS AND RESULTS} \label{section:expe}

\subsection{Experimental Setup}


In this study, we selected the car-following scenario as the experimental focus, as it constitutes a significant portion of driving behavior and effectively reflects diverse driving styles, making it well-suited for validating our method's effectiveness. The experiment utilized the HighD dataset\cite{krajewski2018highd}, which was collected by a 4K resolution camera mounted on a drone flying over a roadway. The data extraction and kinematic model for the car-following behavior were based on FollowNet\cite{chen2023follownet}. To ensure the objectivity of the Statistical Evaluation module, 15\% data was reserved as a test set.

Additionally, to comprehensively assess the impact of various user commands, we followed the approach outlined in\cite{cuiperson} and categorized user commands into three levels based on their directness, as detailed in TABLE \ref{tab:levels}. This classification enables a more precise analysis of how different user commands influence the customization of driving styles. We adopted \textit{GPT-4o-2024-08-06}\footnote{\url{https://platform.openai.com/docs/models}} as the base model and the temperature was set to 0.3 for more deterministic text generation. We set $k=3$ in selecting relevant styles and $n=2$ in Statistical Evaluation.

\begin{table}[]
\centering
\caption{Levels of User Command}
\label{tab:levels}
\begin{tabular}{lll}
\toprule
Level & Linguistic Category & Example Commands \\ \midrule
Level I & Direct & Drive aggressively. \\ \midrule
Level II & Indirect with strong hints & You are braking too harsh. \\ \midrule
Level III & Indirect with mild hints & I am going to be late. \\ \bottomrule
\end{tabular}
\end{table}

\subsection{Customizing Driving Style}

First, we evaluated whether the driving styles generated by Words2Wheels align with the specified user commands. Prior to testing, we populated the Driving Style Database with 8 initial styles. Some of these were data-driven reward functions, derived using the clustering plus IRL approach outlined in\cite{zhao2022personalized}, while others were human-designed reward functions based on the methodologies from\cite{zhu2020safe,zhu2018human,abouelazm2024review}. For generating Style Policy from Style Reward, we employed Proximal Policy Optimization\cite{schulman2017proximal}, a commonly used RL algorithm, for training. Since RL training can be unstable at times, we ran the training 5 times using different seeds, selecting the Style Policy with the highest reward.

\begin{figure}[h]
    \centering
    \includegraphics[width=0.94\linewidth]{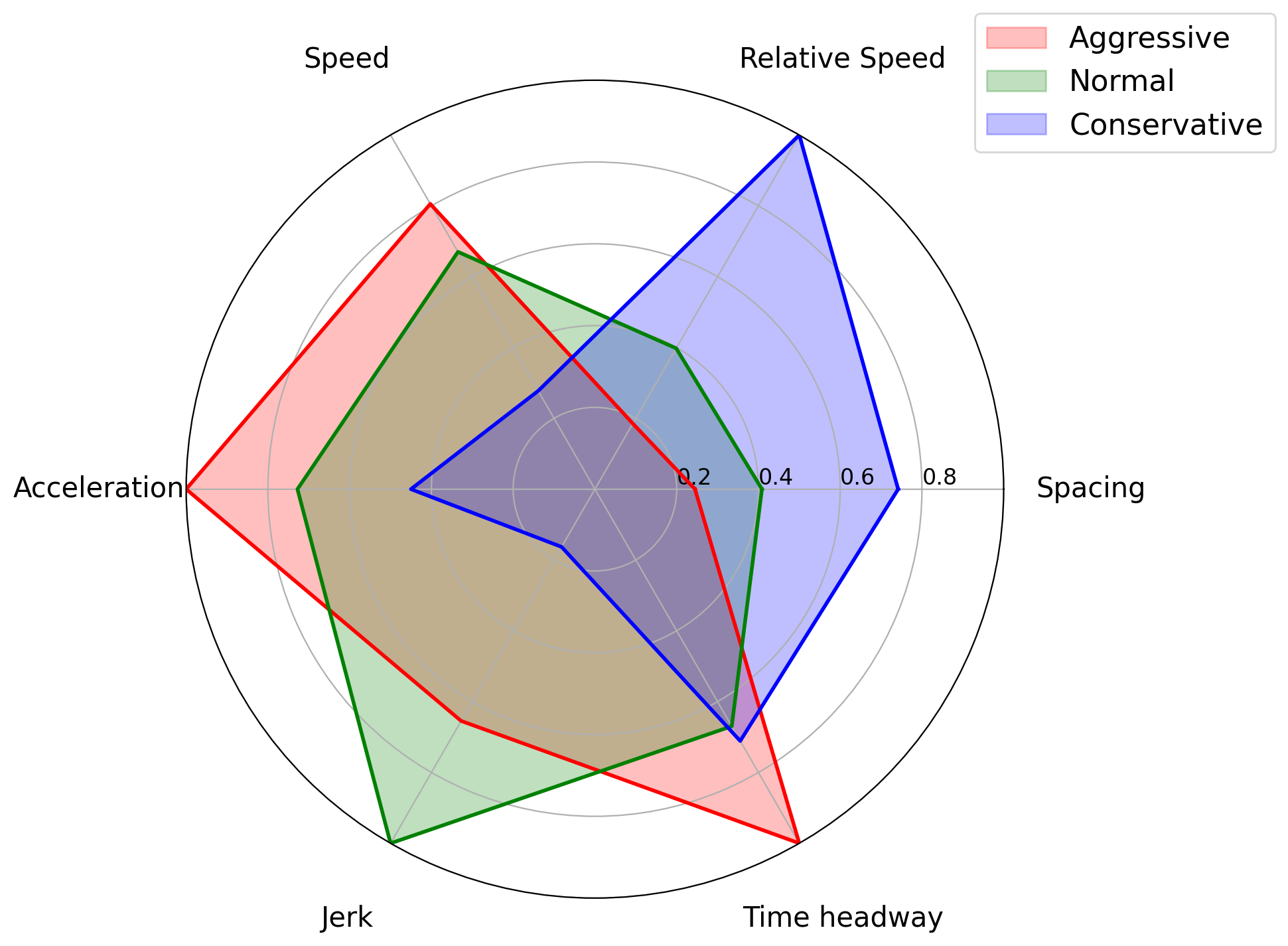}
    \caption{Comparison of customized policies over style-aware metrics.}
    \label{fig:expe_basic}
\end{figure}

We began by testing three specific user commands: \textit{``Drive aggressively / normally / conservatively.''} To ensure the robustness of the results, we ran each command 5 times and averaged the outcomes, as shown in Fig. \ref{fig:expe_basic}. In the repeated experiments, sometimes the newly generated style is selected as the best suited, and sometimes the existing style is. To make the data easier to compare, we normalized it using the 10th and 90th percentiles of the natural driving behavior from the test set. The aggressive style showed higher speeds and greater acceleration, while the conservative style maintained a larger gap and relative speed (leading vehicle minus following vehicle), which aligned well with expectations. However, the aggressive style had a larger time headway, and the normal style showed higher jerk, possibly because the LLM did not prioritize these metrics. Overall, Words2Wheels demonstrated a strong ability to adapt to different driving styles based on user commands.

\subsection{Generation Capability}

In this experiment, we tested Words2Wheels' ability to generate driving styles by only providing styles that did not match the user commands. For example, when the user command indicated an \textit{aggressive} driving style, we \textit{only} provided \textit{conservative} and \textit{normal} styles, and vice versa. We limited the initial styles to 4 for better demonstration.

\begin{figure}[h]
    \centering
    \includegraphics[width=0.94\linewidth]{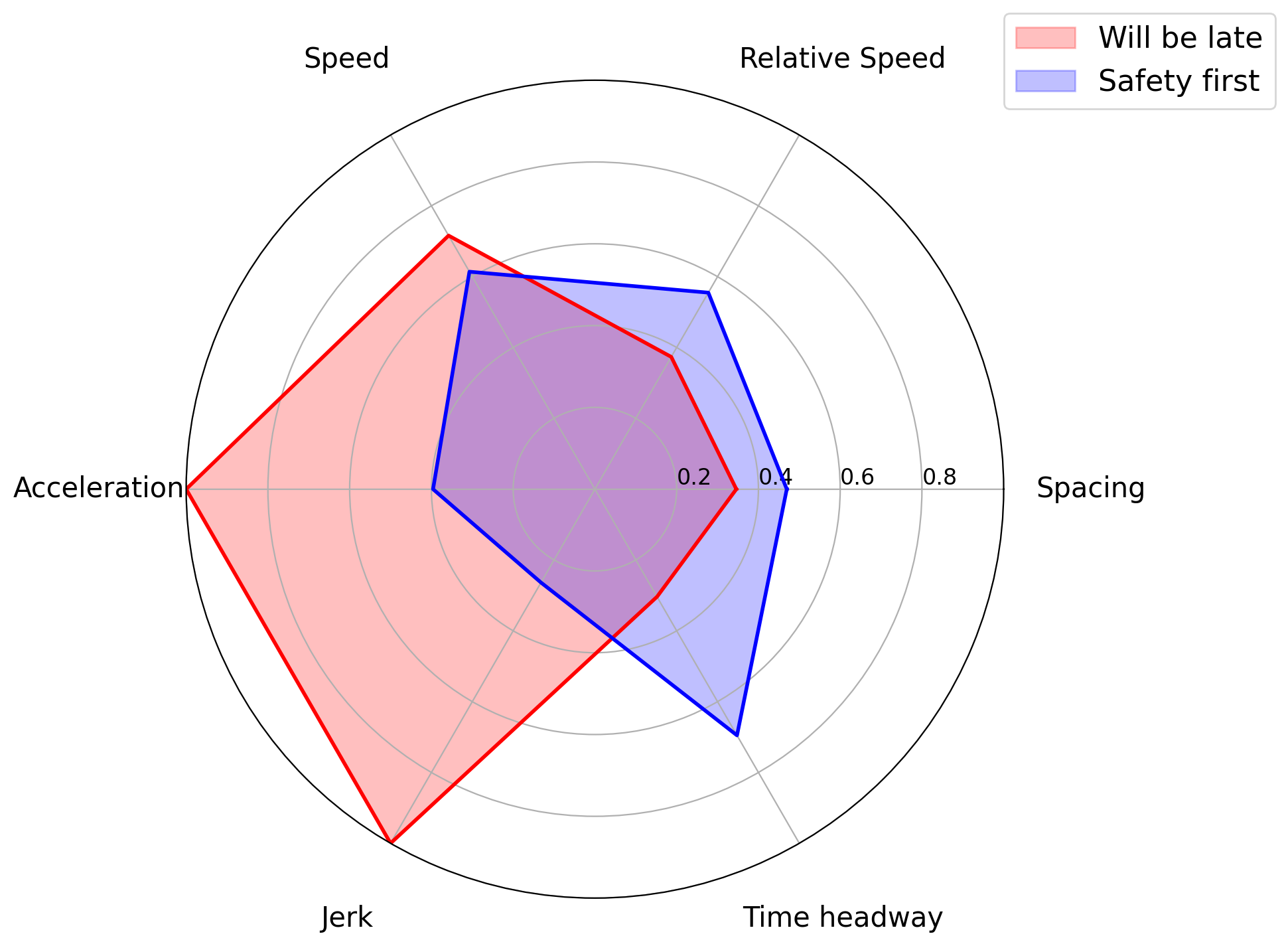}
    \caption{Evaluation of generated policies when no aligned styles provided.}
    \label{fig:expe_generation}
\end{figure}

Results from 5 repeated runs were averaged and are shown in Fig. \ref{fig:expe_generation}. When the input was \textit{``I'm going to be late for the train,''} the generated style exhibited higher speed, acceleration, and jerk. In contrast, when the input was \textit{``Safety first. I have plenty of time,''} the generated style showed greater spacing, time headway, and relative speed. These results demonstrate that Words2Wheels can generate driving styles that align with user commands rather than simply selecting from existing styles. This capability is crucial for generalizing to new user commands and driving scenarios.

\subsection{Human-in-the-Loop Comparisons}

\begin{figure}[h]
    \centering
    \includegraphics[width=0.96\linewidth]{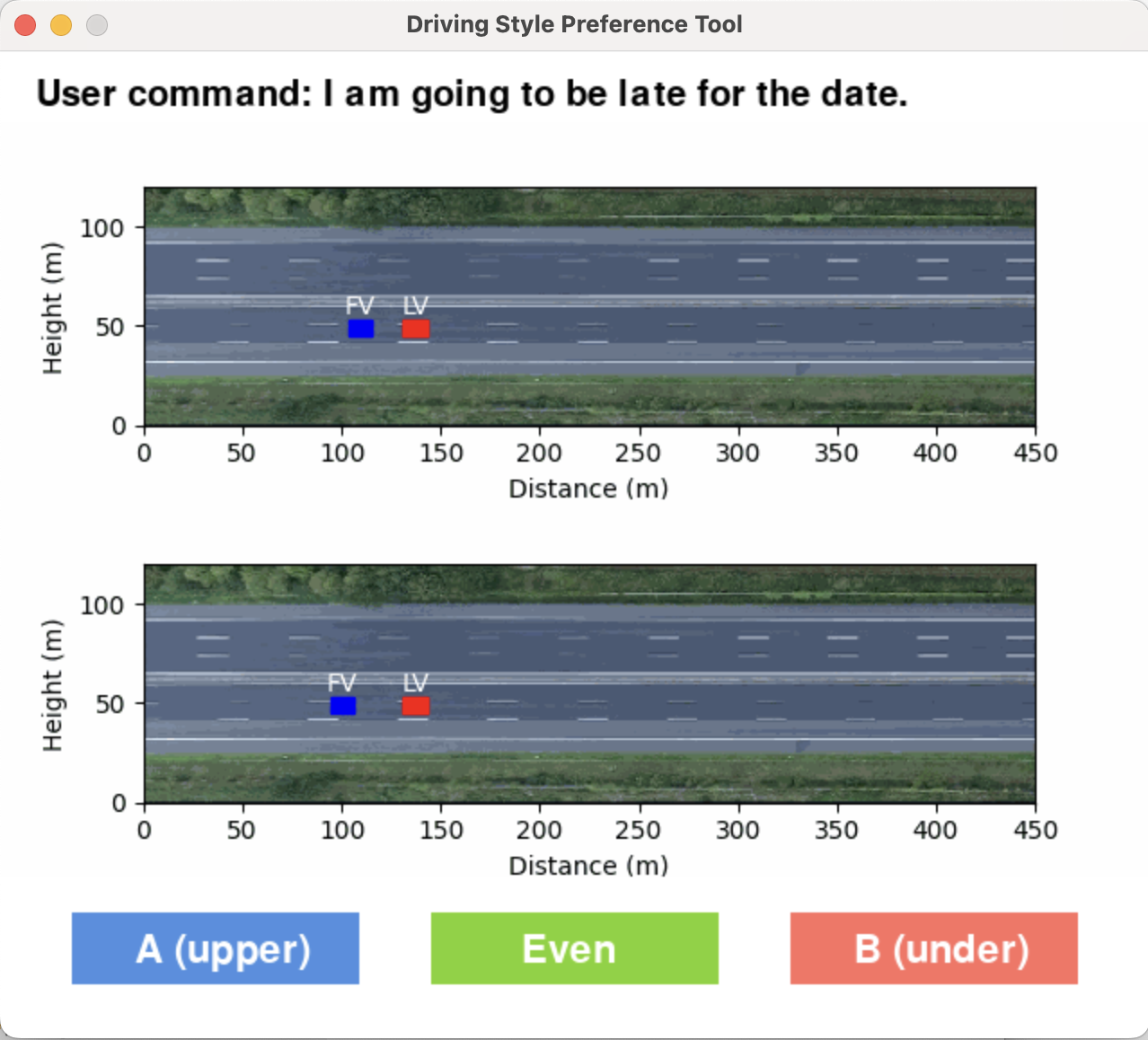}
    \caption{User Interface of the tool for comparing driving style alignment.}
    \label{fig:preference_tool}
\end{figure}

According to research on Reinforcement Learning with Human Feedback\cite{christiano2017deep}, human preferences tend to be more consistent and reliable than direct quantification, particularly in scientific experiments. To leverage this, we developed a visualization tool to compare human judgments on driving style consistency, as illustrated in Fig. \ref{fig:preference_tool}. This tool displays clips of car-following behavior from two different models in the same real-world scenario, allowing users to intuitively select which clip better aligns with the user command or if both seem similar. To eliminate bias, the models are anonymized as A and B.

Referring to prior research\cite{zhao2022personalized,zhou2020modeling}, we benchmarked Words2Wheels against Intelligent Driver Model (IDM)\cite{treiber2000congested}, a widely recognized car-following model. A detailed introduction to IDM and its calibration can be found in\cite{han2023ensemblefollower}.

We recruited 10 volunteers with valid driving license, presenting each with 5 user commands and testing 20 events per command, resulting in a total of 1,000 tested events. The results, summarized in TABLE \ref{human_preferences}, show that in 72\% events, participants judged the driving style generated by Words2Wheels to be a better match to the user command than IDM. In 18.8\% events, both models were considered equally effective, while IDM was preferred in only 9.2\% cases. These findings highlight the capability and effectiveness of Words2Wheels in adapting to user commands.

\begin{table}[]
\centering
\caption{Human Preferences on Driving Style Alignment}
\label{human_preferences}
\begin{tabular}{lllll}
\toprule
Command & \begin{tabular}[c]{@{}l@{}}Prefer\\ Ours\end{tabular} & \begin{tabular}[c]{@{}l@{}}Prefer\\ IDM \end{tabular} & Even & \begin{tabular}[c]{@{}l@{}}Tested\\ events\end{tabular} \\ \midrule
I'm going to be late for the date. & 72\% & 12.5\% & 15.5\% & 200 \\  \midrule
\begin{tabular}[c]{@{}l@{}}I am getting car sick and\\ prefer a smooth ride.\end{tabular} & 64\% & 10.5\% & 25.5\% & 200 \\  \midrule
Drive as fast as you can. & 72.5\% & 8.5\% & 19\% & 200 \\  \midrule
Safety first. I have plenty of time. & 79.5\% & 3.5\% & 17\% & 200 \\  \midrule
\begin{tabular}[c]{@{}l@{}}The cars behind us are honking,\\might be urging us.\end{tabular} & 72\% & 11\% & 17\% & 200 \\ \midrule
Total & 72\% & 9.2\% & 18.8\% & 1000 \\ \bottomrule
\end{tabular}
\end{table}

\subsection{Generalization Capability}

To assess whether Words2Wheels can handle various user commands, we designed and tested 40 commands\footnote{Experimental records available at \url{https://bit.ly/4giYhB7}}, each reflecting a specific driving style, and closely evaluated the LLM's responses for each one. These commands span across three levels of directness. First, we examined whether the LLM correctly interpreted the key factors prioritized by the user, such as efficiency or safety. Then, we verified whether the key metrics selected by the LLM through the Statistical Evaluation module were logically sound. Next, we evaluated the objectivity of the $k$ Style Rewards retrieved from the Driving Style Database. Finally, we reviewed the reasonableness of the generated reward functions.

\begin{table}[]
\centering
\caption{Case Studies across All Command Levels}
\label{case_study}
\begin{tabular}{llllll}
\toprule
\begin{tabular}[c]{@{}l@{}}Command\\ Level\end{tabular} & \begin{tabular}[c]{@{}l@{}}Rational\\ Priorities\end{tabular} & \begin{tabular}[c]{@{}l@{}}Rational\\ Metrics\end{tabular} & \begin{tabular}[c]{@{}l@{}}Rational\\ Retrieval\end{tabular} & \begin{tabular}[c]{@{}l@{}}Rational\\ Generation\end{tabular} & \begin{tabular}[c]{@{}l@{}}Case\\ Count\end{tabular} \\ \midrule
Level I & 100\% & 100\% & 100\% & 90\% & 10 \\ \midrule
Level II & 90\% & 90\% & 90\% & 80\% & 10 \\ \midrule
Level III & 100\% & 95\% & 90\% & 85\% & 20 \\ \midrule
Total & 97.5\% & 95\% & 92.5\% & 85\% & 40 \\ \bottomrule
\end{tabular}
\end{table}

The statistical results of these case studies are presented in TABLE \ref{case_study}. Overall, Words2Wheels performed well in terms of analysis, decision-making, and generation for the vast majority of cases. The system’s handling of Level I commands was slightly more accurate than for Level II and III commands, closely aligning with human expectations. In a few isolated cases, the LLM’s interpretation and processing were off, such as in Case 19, where the LLM misinterpreted the user command \textit{``You are allowing a large gap with the vehicle in front,''} and in Case 26, where the LLM intentionally selected the Style Reward with the highest jerk value as a counterexample. Another recurring issue was that when acceleration and jerk values were negative, the LLM’s understanding conflicted with human intuition. Despite these occasional misinterpretations, the case studies still demonstrated the strong overall capabilities of Words2Wheels.

\subsection{Fuzzy Memory}

To assess the fuzzy memory functionality of Words2Wheels, we conducted 3 experimental sets. In each set, we input a predefined memory into the Driving Style Database and then tested the system with 5 similar inputs to see if it could recall the corresponding memory. For instance, we entered the phrase \textit{``I'm going to be late for the train''} as a historical command linked to a specific driving style in the database. We then tested whether the system could directly match this style when given similar inputs, such as \textit{``I'm going to be late for the plane.''}

The results, shown in TABLE \ref{table:fuzzy_memory}, summarize the system’s ability to recall associated driving styles. With the exception of one case involving \textit{``It’s snowing and visibility has decreased''}, all other test cases were successfully recalled. The system demonstrated a strong ability to recognize and match fuzzy inputs to existing memories. This capability enhances the adaptability of the driving models, enabling the system to respond effectively to a broader range of natural language instructions by recalling similar previous commands.

\begin{table}[]
\centering
\caption{Testing Fuzzy Memory Function with Similar Commands}
\label{table:fuzzy_memory}
\begin{tabular}{ll}
\toprule
Memory & Similar Command (if successfully recall) \\ \midrule
\begin{tabular}[c]{@{}l@{}}I’m going to be\\ late for \textbf{\{the train\}}.\end{tabular} & \begin{tabular}[c]{@{}l@{}}the plane (\textbf{yes}), the party (\textbf{yes}),\\ the game (\textbf{yes}), work (\textbf{yes}),\\ picking up my son from school (\textbf{yes})\end{tabular} \\ \midrule
\begin{tabular}[c]{@{}l@{}}\textbf{\{It's getting dark\}} and\\ visibility has decreased.\end{tabular} & \begin{tabular}[c]{@{}l@{}}It's raining (\textbf{yes}), It's a hazy day (\textbf{yes}),\\ It's foggy outside (\textbf{yes}), It’s snowing (\textbf{no}),\\ The street lights are broken (\textbf{yes})\end{tabular} \\ \midrule
\begin{tabular}[c]{@{}l@{}}I am \textbf{\{dizzy\}} and\\ prefer a smooth ride.\end{tabular} & \begin{tabular}[c]{@{}l@{}}getting car sick (\textbf{yes}), watching movie (\textbf{yes}),\\ having breakfast (\textbf{yes}), not in a hurry (\textbf{yes}),\\ having some food (\textbf{yes})\end{tabular} \\ \bottomrule
\end{tabular}
\end{table}


\section{CONCLUSION AND DISCUSSION} \label{section:conclusion}

This paper presents Words2Wheels, a framework that customizes driving policies based on natural language user commands. By utilizing a Style-Customized Reward Function and a Driving Style Database, the system effectively aligns user preferences with driving behavior. Experiments showed Words2Wheels' ability to generate and adapt driving styles, highlighting its potential for improving user experience in autonomous vehicles through style customization.

Future work will focus on expanding the framework to more complex driving scenarios beyond car-following, refining the Statistical Evaluation module for finer tuning, and incorporating a Feedback and Reflection module\cite{han2024generating} to improve reward functions and enhance system reliability. These developments aim to further enhance Words2Wheels’ adaptability and precision in style-aware driving.

\section*{Acknowledgement}

This study utilizes GPT-4o-2024-08-06 for analysis and code generation in the proposed Words2Wheels framework.

\bibliographystyle{IEEEtran}
\bibliography{refference}

\end{document}